# Effect of Word Embedding Variable Parameters on Arabic Sentiment Analysis Performance

**A. Alnawas [1,2], N. ARICI [1]**

[1] Department of Computer Engineering, Faculty of Technology, Gazi University, 06500 Ankara, Turkey.

[2] Nasiriyah Technical Institute, Southern Technical University, Iraq.

**ABSTRACT**

Social media such as Twitter, Facebook, etc. has led to a generated growing number of comments that contains user's opinions. Sentiment analysis research deals with these comments to extract opinions which are positive or negative. Arabic language is a rich morphological language; thus, classical techniques of English sentiment analysis cannot be used for Arabic. Word embedding technique can be considered as one of successful methods to gaping the morphological problem of Arabic. Many works have been done for Arabic sentiment analysis based on word embedding, but there is no study focused on variable parameters. This study will discuss three parameters (Window size, Dimension of vector and Negative Sample) for Arabic sentiment analysis using DBOW and DMPV architectures. A large corpus of previous works generated to learn word representations and extract features. Four binary classifiers (Logistic Regression, Decision Tree, Support Vector Machine and Naive Bayes) are used to detect sentiment. The performance of classifiers evaluated based on; Precision, Recall and F1-score.

**Corresponding Author:**

N. ARICI

Email: nursal@gazi.edu.tr

Telephone number: 0090312 202 85 73

**ORCID of the authors:**

A. Alnawas: 0000-0001-9181-9377

N. ARICI : 0000-0002-4505-1341

# INTRODUCTION

Over the last few years, the Sentiment Analysis (SA) or opinion mining has been heavily used by researchers as it helps companies and organizations to enhance their products/services (Duwairi et al. 2015). SA is defined as the tasks of finding the opinions of authors about specific topics whether it is a positive or a negative sentiment. SA involves a combination of natural language processing (NLP) and text mining. Many of the studies are dealing with SA for English language. However, there are limited numbers of studies about SA in Arabic (Alnawas and Arıcı 2018). Arabic language needs the complexity of NLP tasks. Many aspects make this complex like; morphology, dialects, orthography, short vowels and word order. For example, one letter in Arabic can be written in many forms. Such as, the letter Hamza (ء) which can be written in four different forms (أ ، ؤ ، ئ ، ء). To overcome the complexity of Arabic, Word embedding or word distributing approach is used to improve NLP in SA tasks. Word2vec model has achieved a good result in with SA Arabic language. Word2vec technique is using neural networks to learn vector representations of words. At the end, the words that are in the same mining or contexts will be closer to each other's. Doc2Vec (derived from Word2Vec) is used to represent sentences into sentence vectors. Unlike other techniques of representation, Doc2Vec takes into consideration the words order and also semantics of the words. Doc2vec was proposed in two architectures: Distributed Bag of Words (DPOW) and Paragraph and Distributed Memory Model of Paragraph Vectors (DM-PV). DBOW represents the input document as a special symbol. DM-PV introduces an additional document symbol to multiple target words. That mines the concatenated vector contain the document symbol and target words (Mikolov et al. 2013b).

In this paper, Doc2Vec model used for Arabic SA and compare four machine learning methods performances that are Logistic Regression (LR), Decision Tree (DT), Support Vector Machine (SVM) and Naive Bayes (NB). Furthermore, the study shows the effect of variable parameters; Windows Size (W), Dimension (D) and Negative Samples (NS) on classifiers' performance.

# RELATED WORKS

There is limited works used word embedding for Arabic SA task. Dahou et al. (2016) used 3.4 billion words to build a word embedding using convolutional neural network. Al-Azani and El-Alfy (2017) used Syria Tweets dataset to experiment highly imbalanced SA datasets. In this study, word embedding with SMOTE approach proposed to handle problem of imbalanced dataset. Zahran et al. (2015) used 5.8 billion from multi different resources. They used different techniques to build word vector in space representations. The Study of Al-Sallab et al. (2017) proposed A Recursive Deep Learning Model for Opinion Mining in Arabic (AROMA) to handle morphological complexity and the lack of opinion resources for Arabic using Recursive Auto Encoder (RAE) model. The Study of Altowayan and Tao (2016) used Word2Vec tool to compute continuous vector representations of Arabic words. For this task, a large Arabic corpus built from different resources. The generated vectors are experimented on six different binary classifiers. Their approach achieved a slightly better accuracy compared to other methods in literature.

# METHODOLOGY

The main goal of this work is to study the effects of Doc2Vec variable parameters on Arabic SA performance. Figure 1 shows the main steps in our methodology.

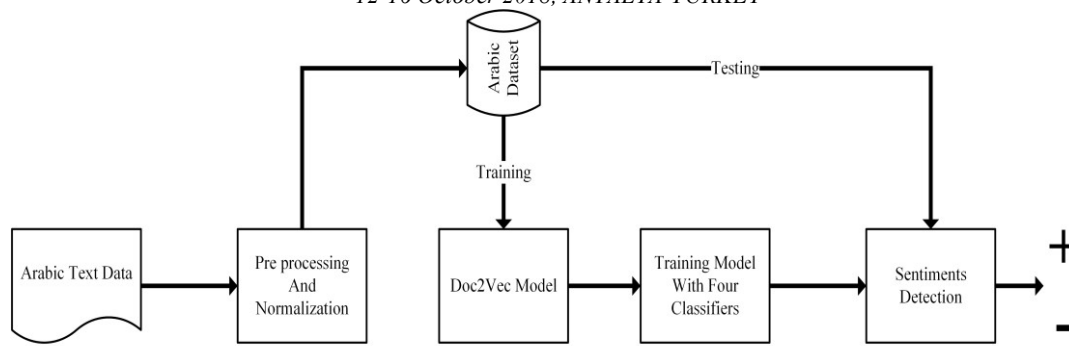

**Fig. 1** General framework

Proposed methodology contains three steps; pre-processing and normalization, represent each comment as vector using Doc2Vec model and feed concatenation vector to the classifiers.

In the pre-processing and normalizing phase, non-Arabic letters, numbers, stop words, punctuations and noise (short vowels) will be deleted. In the second step, every comment will be represented as vector in space with unique sentence ID and its polarity. After we obtained the vectors representing of each comment, four of classifier algorithms applied on those in third step. The last step will predict unclassified comments based on generated model from the previous step.

# EXPERIMENTS

## Datasets

We conducted our experiments using large Arabic multi-domain resources for sentiment analysis corpus that are built by study of (ElSahar and El-Beltagy 2015). The datasets consist of 33K annotated reviews that are collected from different websites for movies, hotels, restaurants and products. We compiled all of the reviews together and performed pre-processing and normalization steps to the combined corpus.

**Table 1.** Summary of Dataset Statistics.

| Dataset | Word Count | Unique Word Count | Positive Reviews | Negative Reviews |
|---|---|---|---|---|
| HTL | 894848 | 80454 | 10049 | 2470 |
| RES | 295790 | 43022 | 7568 | 2513 |
| MOV | 154978 | 36016 | 399 | 135 |
| PROD | 35241 | 9975 | 2759 | 786 |

## Pre-processing and normalization

Before generating the word embedding, the formatting of sentences and words is necessary. The datasets were carefully prepared, we noticed some irregularities in the texts such as the punctuation marks, repetition of letters in the word and some non-Arabic characters. By using Python's Natural Language Toolkit (NLTK), we performed pre-processing on the datasets.

## Doc2Vec model generation

The experiments are conducted with several parameters settings. We have used corpus of (Abu El-Khair 2016) to train Doc2Vec model. To test the performance of the classifiers we apply the generator embedding model with each setting of the parameters. The parameters are set as suggested in the literature, window sizes of 1, 5 and 10 (Bansal et al. 2014; Godin et al. 2015), Dimensions of 100, 300 and 500 (Mikolov et al. 2013b; Tang et al. 2014), and negative samples of 2, 5, 10 and 30 (Mikolov et al. 2013a; Mikolov et al. 2013b). The experiment conducted on both the implementation of Doc2Vec (Le and Mikolov 2014) modified by Linan Qiu (Qiu 2015) and Python's Gensim. To suit the Doc2Vec model with Arabic encoding some modifications are made to achieve training and testing tasks.

## Evaluation protocol

Mostly the standard sentiment classification is a binary classification. We trained generated vectors using four typical machine learning methods: LR, DT, SVM and NB. The classifiers run under the same conditions. Datasets are divided into training and test datasets with 80:20. Performance was quantified using Accuracy (Acc), precision (P), Recall (R) and F1.

## RESULTS

This section highlights the classifiers' performance results based on proposed parameters settings of generated word embedding model. The following figures show the result of our experiments.

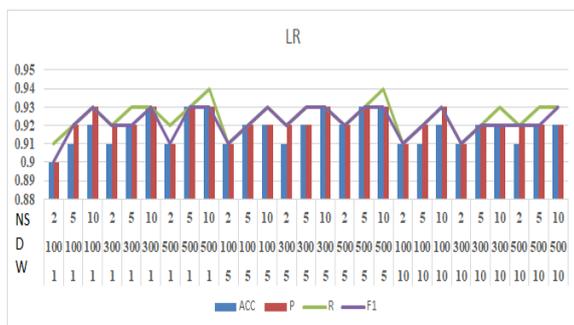
**Fig. 2** LR classifier performance

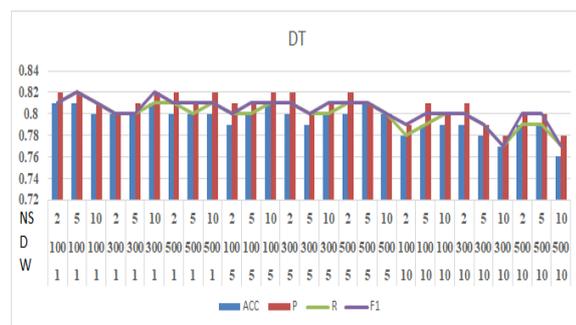
**Fig. 3** DT classifier performance.

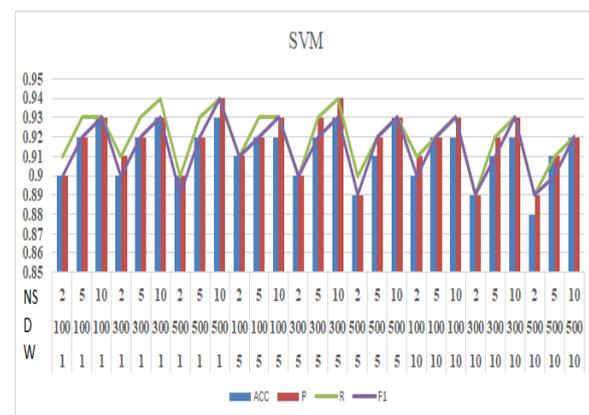
**Fig. 4** SVM classifier performance.

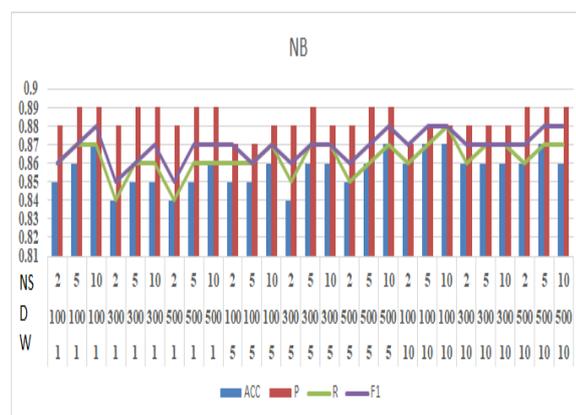
**Fig. 5** NB classifier performance.

Figures 2, 3, 4 and 5 represent the performance for four classifiers based on PV-DM architecture. It is clear that the better performance is always with higher Dimension and Negative Samples. The best Acc equals to 0.93 for both LR and SVM classifiers.

Figures 6, 7, 8 and 9 show the performance for four classifiers based on PV-DBOW architecture. The results show that higher Dimension and Negative Samples give slightly better performance. The best Acc equals to 0.94 for both LR classifier.

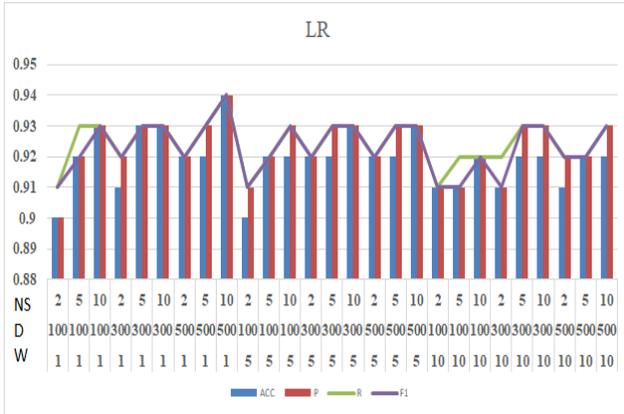

**Fig. 6** LR classifier performance.

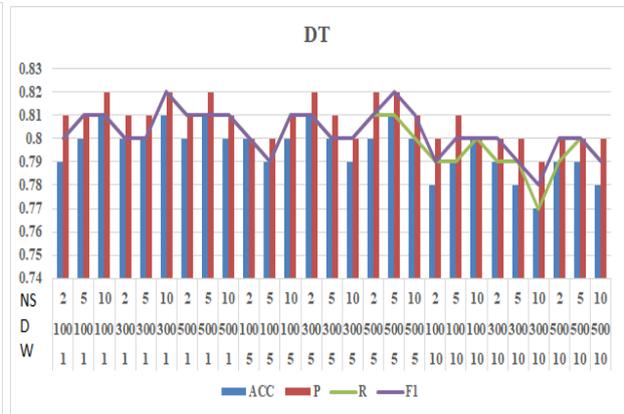

**Fig. 7** DT classifier performance.

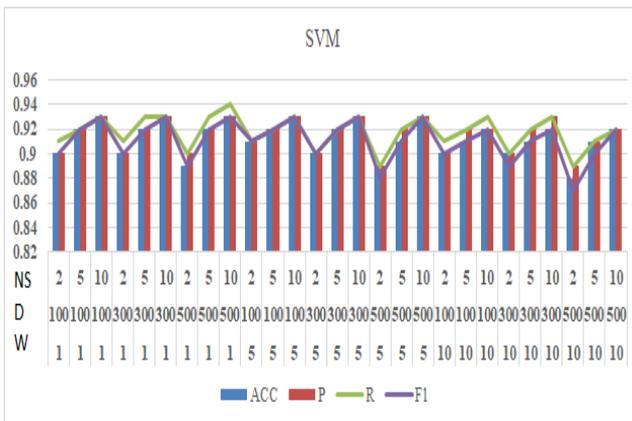

**Fig. 8** SVM classifier performance.

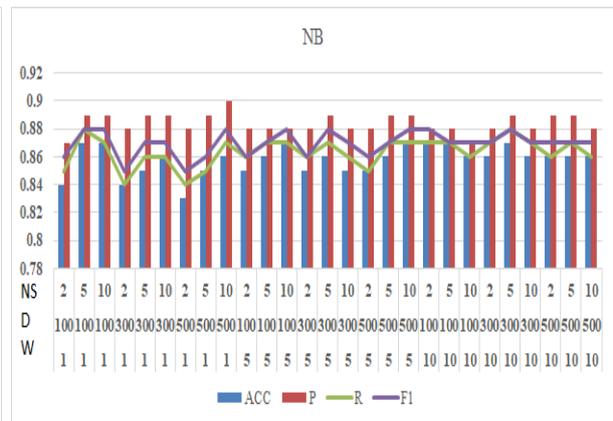

**Fig. 9** NB classifier performance.

The experiments show a smaller context window size gives a better performance, larger Dimension and Negative Samples give slightly better performance. The best setup of parameters may differ from a task to another based on background corpus, balance of polarity and classifiers that used.

## CONCLUSION

In this paper, distributed representations of documents are modelled for Arabic Sentiment Analysis. We study the effect of various variable parameters setup of Doc2Vec model (Window Size, Dimension of vector and Negative Sample) on the four of machine learning methods: LR, DT, SVM and NB. Two architectures of Doc2vec are applied; DPOW and DM-PV. In conclusion, Doc2Vec with both large Dimension and Negative Samples are better to achieve high effectiveness with classifiers. In contrast, small context window is preferable to attain high effectiveness.

**Acknowledgements** The authors would like to thank all who supported this work.

# REFERENCES

Abu El-Khair I (2016) 1.5 billion words Arabic Corpus arXiv preprint arXiv:161104033 doi:2016arXiv161104033A

Al-Azani S, El-Alfy E-SM (2017) Using word embedding and ensemble learning for highly imbalanced data sentiment analysis in short arabic text. In: Proceedings of The 8th International Conference on Ambient Systems, Networks and Technologies, ANT Madeira, Portugal, 2017. Procedia Computer Science, pp 359-366. doi:10.1016/j.procs.2017.05.365

Al-Sallab A, Baly R, Hajj H, Shaban KB, El-Hajj W, Badaro G (2017) Aroma: A recursive deep learning model for opinion mining in arabic as a low resource language ACM Transactions on Asian and Low-Resource Language Information Processing (TALLIP) 16:25 doi:10.1145/3086575

Alnawas A, Arıcı N (2018) The Corpus Based Approach to Sentiment Analysis in Modern Standard Arabic and Arabic Dialects: A Literature Review Journal of Polytechnic 21:461-470 doi:10.2339/politeknik.403975

Altowayan AA, Tao L (2016) Word embeddings for Arabic sentiment analysis. In: Proceedings of IEEE International Conference on Big Data, Washington, DC, USA, 2016. IEEE, pp 3820-3825. doi:10.1109/BigData.2016.7841054

Bansal M, Gimpel K, Livescu K (2014 ) Tailoring continuous word representations for dependency parsing. In: Proceedings of the 52nd Annual Meeting of the Association for Computational Linguistics (Volume 2: Short Papers), Baltimore, Maryland, USA, 2014. pp 809-815. doi:10.3115/v1/P14-2131

Dahou A, Xiong S, Zhou J, Haddoud MH, Duan P (2016) Word embeddings and convolutional neural network for arabic sentiment classification. In: Proceedings of the 26th International Conference on Computational Linguistics: Technical Papers, Osaka,Japan, 2016. pp 2418-2427

Duwairi R, Ahmed NA, Al-Rifai SY (2015) Detecting sentiment embedded in Arabic social media–a lexicon-based approach Journal of Intelligent & Fuzzy Systems 29:107-117 doi:10.3233/IFS-151574

ElSahar H, El-Beltagy SR (2015) Building large arabic multi-domain resources for sentiment analysis. In: Proceedings of 16th International Conference on Intelligent Text Processing and Computational Linguistics, Cairo, Egypt, 2015. Springer, pp 23-34. doi:10.1007/978-3-319-18117-2_2

Godin F, Vandersmissen B, De Neve W, Van de Walle R (2015) Multimedia Lab @ ACL WNUT NER Shared Task: Named Entity Recognition for Twitter Microposts using Distributed Word Representations. In: Proceedings of the Workshop on Noisy User-generated Text, Beijing, China, 2015. pp 146-153. doi:10.18653/v1/W15-4322

Le Q, Mikolov T (2014) Distributed representations of sentences and documents. In: Proceedings of International Conference on Machine Learning, Beijing, China, 2014. pp 1188-1196. doi:2014arXiv1405.4053L

Mikolov T, Chen K, Corrado G, Dean J (2013a) Efficient estimation of word representations in vector space arXiv preprint arXiv:13013781 doi:2013arXiv1301.3781M

Mikolov T, Sutskever I, Chen K, Corrado GS, Dean J (2013) Distributed representations of words and phrases and their compositionality. In: Proceedings of Advances in Neural Information Processing Systems 26, Nivada.USA, 2013b. pp 3111-3119. doi:2013arXiv1310.4546M

Qiu L (2015) Sentiment Analysis using Doc2Vec in gensim. https://github.com. https://github.com/linanqiu/word2vec-sentiments. Accessed December 2017

Tang D, Wei F, Yang N, Zhou M, Liu T, Qin B (2014) Learning sentiment-specific word embedding for twitter sentiment classification. In: Proceedings of the 52nd Annual Meeting of the Association for Computational Linguistics, Baltimore, Maryland, USA, 2014. pp 1555-1565. doi:10.3115/v1/P14-1146

Zahran MA, Magooda A, Mahgoub AY, Raafat H, Rashwan M, Atyia A (2015) Word representations in vector space and their applications for arabic. In: Proceedings of the International Conference on Intelligent Text Processing and Computational Linguistics, Cairo, Egypt, 2015. Springer, pp 430-443. doi:10.1007/978-3-319-18111-0_32